\documentclass[conference]{IEEEtran}

\IEEEoverridecommandlockouts
\usepackage{tikz}
\usepackage{textcomp}
\usepackage{hyperref}
\usepackage{lipsum}

\newcommand\copyrighttext{%
  \footnotesize \textcopyright 2021 IEEE.  Personal use of this material is permitted.  Permission from IEEE must be obtained for all other uses, in any current or future media, including reprinting/republishing this material for advertising or promotional purposes, creating new collective works, for resale or redistribution to servers or lists, or reuse of any copyrighted component of this work in other works.}
\newcommand\copyrightnotice{%
\begin{tikzpicture}[remember picture,overlay]
\node[anchor=south,yshift=10pt] at (current page.south) {\fbox{\parbox{\dimexpr\textwidth-\fboxsep-\fboxrule\relax}{\copyrighttext}}};
\end{tikzpicture}%
}
\usepackage{cite}
\usepackage{amsmath,amssymb,amsfonts}

\usepackage[ruled,linesnumbered]{algorithm2e}
\SetAlFnt{\small}
\usepackage{array,mathtools, multirow}
\usepackage{graphicx}
\usepackage{textcomp}

\SetCommentSty{mycommfont}
\usepackage{xcolor}
\def\BibTeX{{\rm B\kern-.05em{\sc i\kern-.025em b}\kern-.08em
    T\kern-.1667em\lower.7ex\hbox{E}\kern-.125emX}}

\begin{document}

\title{Communication and Computation Reduction for Split Learning using Asynchronous Training}
\author{\IEEEauthorblockN{Xing Chen, Jingtao Li and Chaitali Chakrabarti}
\IEEEauthorblockA{
School of Electrical, Computer and Energy Engineering\\
Arizona State University, Tempe, USA \\
xchen382@asu.edu, jingtao1@asu.edu, chaitali@asu.edu}}

\maketitle
\copyrightnotice
\begin{abstract}
Split learning is  a promising privacy-preserving distributed learning scheme that has low computation requirement at the edge device but has the disadvantage of high communication overhead between edge device and server. To reduce the communication overhead, this paper proposes a loss-based asynchronous training scheme that updates the client-side model less frequently and only sends/receives activations/gradients in selected epochs.  To further reduce the communication overhead, the activations/gradients are quantized using 8-bit floating point prior to transmission. An added benefit of the proposed communication reduction method is that the computations at the client side are reduced due to reduction in the number of client model updates. Furthermore, the privacy of the proposed communication reduction based split learning method is almost the same as traditional split learning. Simulation results on VGG11, VGG13 and ResNet18 models on CIFAR-10 show that the communication cost is reduced by 1.64x-106.7x and the computations in the client are reduced by 2.86x-32.1x when the accuracy degradation is less than 0.5\% for the single-client case.
For 5 and 10-client cases, the communication cost reduction is 11.9x and 11.3x on VGG11 for 0.5\% loss in accuracy.

\end{abstract}


\section{Introduction}
Data security has become a big concern in traditional Deep Neural Network (DNN) training where raw data at edge are collected and processed by a central server. Even if the server is honest, data can be leaked through membership inference and model inversion attacks \cite{shokri2017membership, zhang2020secret}.  To address data privacy, cryptographic approaches such as Multi-party computation \cite{wagh2019securenn} and Homomorphic Encryption \cite{nandakumar2019towards} have been proposed. These techniques are  computationally intensive and not suitable for  edge devices.

Techniques such as federated learning \cite{konevcny2016federated} and split learning \cite{gupta2018distributed} preserve the privacy in distributed learning and are more resource-friendly compared to cryptographic approaches.
Federated learning aggregates model parameter updates from clients in a central server.
It requires  all clients to be able to train the entire model periodically. However, clients usually run on edge devices, which have limited computation resources, making it hard to support federated learning. Furthermore, the server in federated learning has full knowledge of the model, making it a potential security problem\cite{thapa2020splitfed}.

\begin{figure}[!ht]
    \centering
    \includegraphics[width=0.95\linewidth]{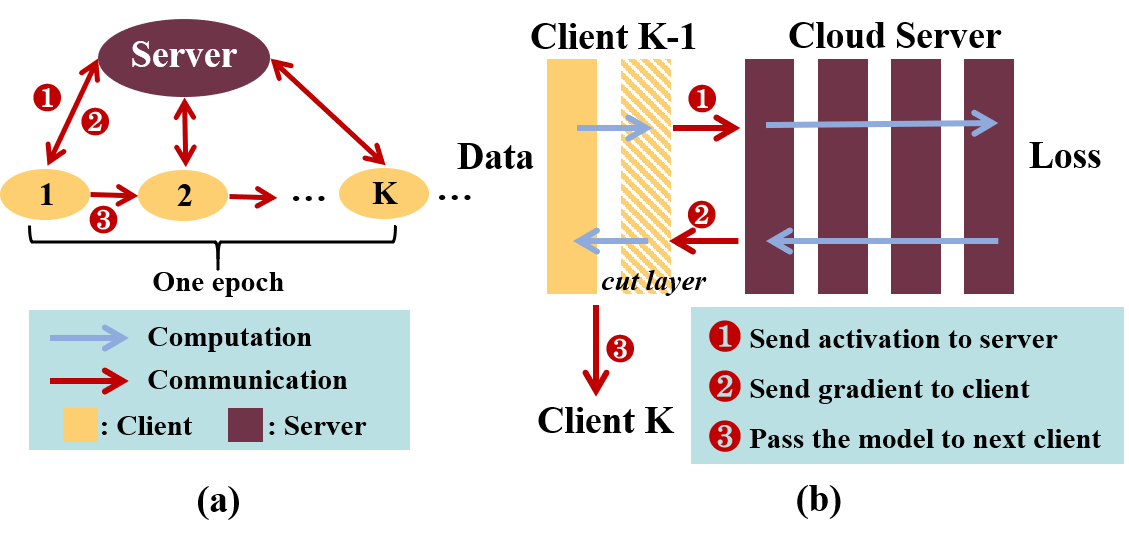}
    
    %
    
    \caption{Overview of split learning scheme. (a) Split learning with K clients. (b) Communication and computation at the client level.}
    \label{fig:split_learn}
    
\end{figure}

Split learning\cite{gupta2018distributed}, on the other hand, splits the model into client-side model and server-side model, and the two parts are trained in a distributed way, as shown in Fig.\ref{fig:split_learn}. Each client computes forward propagation till a particular layer, called \textit{cut layer} (slashed yellow rectangle in the figure). The client sends the activation of the cut layer and the labels of its data to the server. The server continues forward propagation on rest of the network followed by backpropagation till the cut layer, and then sends the gradients back to the client.
After a local epoch\footnote{One local epoch of a client corresponds to complete forward and  back-propagation  for all local data of that client.
}, the client passes the latest client-side model parameter to the next client (a.k.a. peer-to-peer mode  \cite{gupta2018distributed}), to synchronize the client-side model across all clients.
In this paper, we ignore the communication of sending the latest model to the next client, since it is negligible compared to that of sending/receiving activation/gradient data.

Split learning has less computational requirement at the edge device since it only needs to process forward/backward propagation of the client-side model\cite{palanisamy2021spliteasy,vepakomma2018split}.
However, the communication overhead linearly increases with the number of training samples. In the extreme case, where the number of edge devices is small and each edge device has to process a large amount of data, communication overhead can be way higher than federated learning \cite{singh2019detailed, gao2020end}.


Recent studies on split learning evaluate convergence \cite{vepakomma2018split}, privacy capability\cite{abuadbba2020can} and implementation on IoT\cite{gao2020end}. An empirical evaluation of split learning in real-world IoT settings in terms of learning performance and device implementation overhead is presented in \cite{gao2020end}. However, none of these works have focused on the reduction of communication overhead in split learning. 

In this paper, we propose a method to reduce the communication overhead associated with split learning to make it more practical. The communication overhead is reduced in two ways: 1) reducing update frequency of the client-side model, and 2) quantization. A loss-based asynchronous training is proposed to reduce the update frequency. Specifically, the client-side model is updated only if the loss drop is greater than a threshold. For cases when the client-side model is not updated, it is unnecessary for the client to send/receive activation/gradient to and from the server. We further  quantize activation/gradient from 32-bit floating point to 8-bit floating point without much accuracy degradation. A search-based quantization scheme is implemented to search for the best combination of exponent bits and bias to quantize the activation/gradient prior to dispatch. We analyze the performance of the proposed approach on VGG11, VGG13 and ResNet18 models on cIFAR-10 dataset for single-client and multi-client cases. We also analyze the privacy of the proposed approach.

This paper makes the following contributions:
\begin{itemize}
    \item  To the best of our knowledge, we are the first to address communication overhead problem in split learning. The proposed method uses a combination of asynchronous training and quantization. Evaluation on the three models for the one-client case  show communication reduction of 1.64x-106.7x and 2.4x-266.7x compared to the baseline scheme for 0.5\% and 1.0\% accuracy loss, respectively.
    \item The proposed scheme requires fewer updates in client-side model, thereby reducing the client's computation drastically. Evaluation of the three models on the one-client case show that the client's computation can be reduced by 2.86x-32.1x and 5.96x-80.3x for 0.5\%  and 1.0\% accuracy loss, respectively.
    \item The proposed loss-based asynchronous training method determines which epoch should update the client-side model. Compared to the case when the epochs to update the client-side model are uniformly spread across the whole training process, the proposed selective update scheme can achieve over 1\% higher accuracy for the same reduction in communication cost.
    
\end{itemize}

\section{Motivation}

The communication overhead of split learning linearly scales with the amount of training data at the client \cite{singh2019detailed}. While split learning has less communication overhead than federated learning \cite{konevcny2016federated} when the data size is small, it is a bottleneck if the data size is large. 
Furthermore, the size of the activations/gradients sent/received to/from the server depends on the location of the cut layer.
Consider a one-client split learning with 30,000 training samples using VGG11. 
When the first 2, 5 and 7 layers are processed at the client side, the size of activations/gradients sent/received by the client are $16\times16\times64$, $8\times8\times256$ and $4\times4\times256$ for every training sample, respectively.

\begin{table}[!ht]
\centering
\caption{Computation and communication time (in minutes) using split learning  with different number of layers at client side. }
    \resizebox{1.0\linewidth}{!}{
\begin{tabular}{c|c|c|c}
\hline
\textbf{VGG11 Number of layers at client-side}     & \textbf{2}  & \textbf{5} & \textbf{7} \\ \hline
\textbf{Communication Time (min)} & 98              & 98    &  24     \\ \hline
\textbf{Computation Time= Client+Server(min)}   & 51=39+12       & 94=84+10     &115=107+8
\\ \hline
\end{tabular}
}
\label{tab:spl_comm_comp}
\end{table}

Table \ref{tab:spl_comm_comp} shows the computation time and communication time breakdown of a system where the client uses an Intel-i7 CPU and the server uses a NVIDIA 1080Ti GPU and the communication speed between client and server is 1Gbps. 
We see that the communication cost is significant when the number of training samples is large.
We also see that this cost decreases compared to the computation time when the number of layers at the client-side increases. The communication cost is dominant up to the case when 5 layers are processed at the client-side and so we consider the cut layer to be less than or equal to 5 in Section IV.

\section{Proposed Method}
\subsection{Loss-based Aasynchronous training}

In this paper we describe an asynchronous training approach for split learning that reduces the number of client-side model updates  to achieve significant communication reduction with minimal accuracy drop. Our method is inspired by federated learning \cite{konevcny2016federated,sattler2019sparse} which achieves minimal loss in accuracy even though the weight updates in different clients are sent to the central server asynchronously and the model is updated at the server using stale information. 
In the proposed loss-based asynchronous training scheme, the server-side model is trained as usual while the client-side model does not update as frequently. In fact the client-side model only updates when the loss difference with that of the last update is larger than a pre-defined loss threshold $l_{thred}$.

  \begin{figure}[!h]
     \centering
         \includegraphics[width=0.95\linewidth]{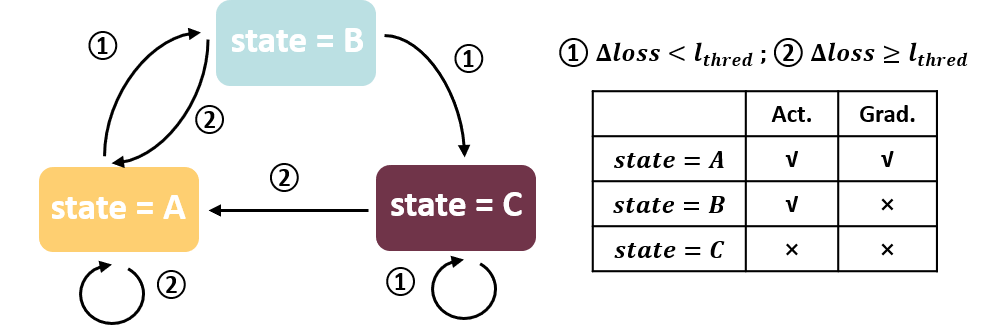}
         
        \caption{State diagram describing the data transfer between clients and server.  $state=A$ sends/receives activation/gradient to/from server, $state=B$ only sends activation to server and $state=C$ has no communication with server. The state transition depends on the change of loss, $\Delta loss$.}
        \label{fig:State Diagram}
\end{figure}
 
 In the proposed scheme, we define $state$ to represent whether the activation should be sent from clients to server and gradient from server to clients in the current epoch. 
The state diagram is shown in Fig.\ref{fig:State Diagram}. The state is updated every epoch based on whether the changes of loss exceed the given loss-threshold $l_{thred}$. When $state=A$, the communication is as in traditional training, where both activation and gradient are transferred to and from the server. When $state=B$, the activation is sent to the server but the server does not send the gradient to the clients.
When $state=C$, there is no communication between clients and server.
The server uses the previous activation of cut layer for its forward and backward computations.

The following is an example with one-client. If the client-side model is updated in epoch $n$, and does not update in epoch $n+1$, then
in epoch $n+1$, the client  does not receive gradient from the server but has to send activation to the server since the client-side model was updated in the last epoch ($state=B$). If the client-side model still does not update in epoch $n+2$, then the activation in epoch $n+2$ is exactly identical to that of epoch $n+1$, so the activation is not sent to the server, and the communication due to activation is also saved ($state=C$).

 
\begin{algorithm}[!htbp]
\SetNoFillComment
\SetAlgoLined
\SetKwInOut{Input}{Input}
\Input{Number of batches $num\_batch$, number of clients $K$, number of epochs $num\_epoch$, loss function $f(\cdot)$, dataset for clients $(x,y)$, model is split to client-side model $W^C(\cdot)$ and server-side model $W^S(\cdot)$, and loss threshold $l_{thred}$.}
\SetKwFunction{FMain}{Asynchronous\_Split\_Learning}
    \SetKwProg{Fn}{def}{:}{}
    \Fn{\FMain{}}{
$state\leftarrow A$ \tcp{set initial state} 
$loss\leftarrow 0$\;
$W^C(\cdot), W^S(\cdot) \leftarrow split (model)$\;

\For{$epoch \gets 1$ \textbf{to} $num\_epoch$} {
    $total\_loss\leftarrow 0$\;
    \For{$client \gets 1$ \textbf{to} $K$} {
        \For{$batch \gets 1$ \textbf{to} $num\_batch$} {
         \tcp{Forward}
        $loss \leftarrow split\_forward()$\;
        $total\_loss \leftarrow total\_loss + loss$\;
         \tcp{Backward}
        $split\_backward()$\;
        }
    send $W^C(\cdot)$ to next client\;
    }
    
    $state\leftarrow update\_state()$ \tcp{Update state}
}

\SetKwFunction{FMain}{split\_forward}
    \SetKwProg{Fn}{def}{:}{}
    \Fn{\FMain{$W^C(\cdot), W^S(\cdot)$, state, (x,y), $f(\cdot)$}}{
       \eIf{$state = C$}{
           $act\leftarrow act^*, y\leftarrow y^*$\;
           }{
           $act\leftarrow W^C(x)$\;
           Send $act, y$ to server\;
           $act^* \leftarrow act, y^* \leftarrow y$\;
           }
            $y' \leftarrow W^S(act)$\;
            $loss\leftarrow f(y,y')$\;
\textbf{return} $ loss$\ 
}
\textbf{end}\

\SetKwFunction{FMain}{split\_backward}
\SetKwProg{Fn}{def}{:}{}
\Fn{\FMain{$W^C(\cdot), W^S(\cdot)$, state,loss}}{
       $grad\leftarrow server\_backward(loss)$\;
       Update $W^S$\;
   \If{$state = A$}{
       Send $grad$ to clients\;
       $client\_backward(grad)$\;
       Update $W^C$\;
       }
}
\textbf{end}\

\SetKwFunction{FMain}{update\_state}
    \SetKwProg{Fn}{def}{:}{}
    \Fn{\FMain{state,total\_loss,$l_{thred}$}}{
   \If{$state=A$}{
   $last\_update\_loss \leftarrow\frac{total\_loss}{num\_batch\times K}$\;
   }
   $\Delta loss \leftarrow last\_update\_loss-\frac{total\_loss}{num\_batch\times K}$\;
   \eIf{$\Delta loss\geqslant l_{thred}$}{
   $state\leftarrow A$\;
   }{
   \eIf{$state=A$}{
   $state\leftarrow B$\;}{
   $state\leftarrow C$\;}
   }

\textbf{return} $state$\ 
}
\textbf{end}\

}

\caption{Round-Robin Split Learning with Loss-based Asynchronous Training}
\label{alg:algorithm1}
\end{algorithm}

The detailed algorithm is given in \textbf{Algorithm\ref{alg:algorithm1}}. The system is set as $state=A$ by server in the first epoch (line 1). In every epoch all clients process in a round-robin fashion (line 6-13). At the end of each epoch, the state is updated by server using $update\_state()$. 

During the forward step $split\_forward()$, if $state = C$, server only needs to read the previous stored activation (line 18). If $state \neq C$, the client computes the cut layer activation and sends it to the server (line 20-22).
During backward $split\_backward()$, only when $state=A$, the gradient from server is sent back to client and client-side model is updated (line 32-34), otherwise, the client-side model is not updated.

In $update\_state()$, if $state=A$, server computes the average loss of the epoch and records it in $last\_update\_loss$ as the loss of the latest update (line 39). Otherwise, server compares the average loss of this epoch with $last\_update\_loss$ (line 4). If the change in loss reaches the set loss threshold $l_{thred}$, the client-side model will update in the next epoch (line 43-50).

\subsection{Search-based Quantization}

 Quantization is used widely in DNN model compression\cite{sun2020ultra,sun2019hybrid,courbariaux2016binarized}. Since fixed point representation cannot represent dynamic range of activation and gradient well, we adopt the 8-bit floating point format\cite{sun2019hybrid}. The activation/gradients are quantized using 8-bit floating point instead of the original 32-bits before being sent to server/clients, to further reduce the communication. Floating point representation consists of 3 parts: sign bit, exponent bits $ebit$, mantissa bits $mbit$. We also introduce exponent bias $bias$\cite{sun2019hybrid} to scale the dynamic range; $bias$ is shared by all values. The absolute dynamic range is from  $[2^{-mbit-bias},\frac{2^{mbit+1}-1}{2^{2^{mbit}}}\times 2^{2^{ebit}-1-bias}]$. Reducing from 32-bit to 8-bit causes two problems: 1) precision loss of values in dynamic range and 2) clipping of overflow and underflow values outside the dynamic range. It has been reported in  \cite{sun2019hybrid,courbariaux2016binarized}  clipped values seriously affect the accuracy and so we ensure that the proportion of clipped values is less than 1\% in the proposed approach.

Due to the variation in the range of activation and gradient across epochs, it is hard to fix the number of exponents bits and exponent bias across epochs while keeping high enough precision. Hence, a search-based quantization method is proposed to search for the best combination of exponent bits and bias so that the proportion of clipped values is less than 1\%. The detailed algorithm is shown in \textbf{Algorithm}\ref{algorithm3}. The number of candidate exponent bits is 3, 4, 5 and 6  (line 2). The candidate bias range is computed such that the maximum value of the floating point with bias should not be less than the median of gradient/activation, and the minimum value should not larger than the median of gradient/activation (line 4-5). If the proportion of overflow and underflow value is smaller than $1\%$, the current exponent bit and bias are returned (line 9-11). If no configuration satisfies the clip  (overflow and underflow) requirement after searching, the activation/gradient in the local epoch will not be quantized (line 13). The search process is conducted in the first batch of every local epoch and the chosen configuration is used by all the other batches in the same local epoch. Thus the computation overhead of this method is quite low compared to the forward/backward computations in a batch.


\begin{algorithm}
\SetAlgoLined
\SetKwInOut{Input}{Input}
\SetKwInOut{Output}{Output}
\Input{gradient/activation to be sent $X$}
\Output{exponent bits $ebit$ and exponent bias $bias$}
\tcp{find the median value of the absolute input}
$median=median(|X|)$\; 
\For{$ebit \gets 3$ \textbf{to} $6$}{
compute the max and min positive value with current $ebit$ without $bias$ as $max\_value$ and $min\_value$\;
\tcp{compute the bias range}
$max\_bias=\log_{2} \frac{min\_value}{median}$\; 
$min\_bias=\log_{2} \frac{max\_value}{median}$\;
\For{$bias \gets min\_bias$ \textbf{to} $max\_bias$}{
compute the proportion of overflow and underflow of $X$ with $ebit$ and $bias$ as $overflow$ and $underflow$\;
$clip=overflow+underflow$\;
\If{$clip<1\%$}{
return $ebit$, $bias$;
}}
exit()\;
}

 \caption{Search-based Quantization}
 \label{algorithm3}
\end{algorithm}

\section{Simulation Results}
 In this section, we demonstrate the trade-off between accuracy and communication reduction using the proposed communication reduction method. We present the results for 1 client followed by the multi-client case. We also discuss the effect of the proposed scheme on privacy.
\subsection{Experimental Setting}
We use Pytorch to simulate split learning, and set all gradients of client-side model to 0 if $state \neq A$. We validate our approach on three types of image classification machine learning models for CIFAR10: VGG11, VGG13 and ResNet18. The number of epochs is 200. We set the maximum number of layers in client-side model to be 5 since in that case the communication and computation overhead are comparable. With more layers, the computation overhead increases and the communication overhead is relatively less important, which is consistent with the result in \cite{gao2020end}. The cut layer of different machine learning models are set as follows:
\begin{itemize}
    \item \textbf{VGG11/VGG13}  VGG mainly consists of 3x3 convolution layers (with ReLU) and maxpooling layers. The maxpooling layer directly connects to the convolution layer in the client-side model and is included at the client side.
    We consider \textit{small/large} setting for VGG networks. \textit{Small}: The first convolution layer is in client-side model (so 1 layer for VGG13 and 2 layers for VGG11 since the convolution layer is followed by a maxpooling layer). \textit{Large}: The first three convolution layers are in client-side model (so 5 layers for VGG11 and 4 layers for VGG13).
    
    \item \textbf{ResNet18} ResNet18 includes a standard 3x3 convolution layer and BasicBlock with different sizes. We consider two split settings. \textit{Small}: The first convolution layer is in client-side model. \textit{Large}: The first convolution and the first two BasicBlocks are in client-side model (each consists of two convolution layers and one skip connection).
    
\end{itemize}

\subsection{Experimental Results}
\subsubsection{Accuracy Performance}
\begin{figure}[!t]
    \centering
    \includegraphics[scale = 0.5]{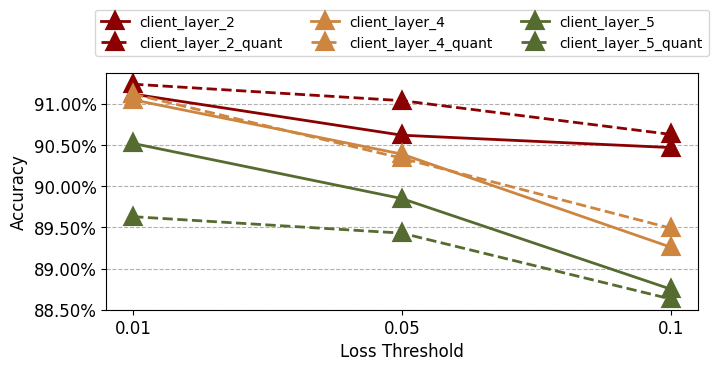}
        \caption{VGG11 Accuracy as a function of loss threshold for different number of client-side layers. Baseline accuracy is 91.5\%.}
    \label{fig:vgg11-accuracy}
    \vspace{-5pt}
\end{figure}

Fig.\ref{fig:vgg11-accuracy} shows how the choice of loss threshold and different number of layers at the client side affect the accuracy of our proposed method. The solid curves correspond to loss-based asynchronous training scheme and the dash curves correspond to asynchronous training along with search-based quantization. The baseline accuracy for the standard split learning with 1-client is 91.5\%. Not surprisingly, when the loss threshold increases, the update frequency of client-side model decreases, resulting in accuracy degradation for both cases. Also, when more layers at client side are updated at lower frequency, the accuracy drop is even more.

For the case when the number of layers in client-side model is 2, and both asynchronous training and quantization are implemented,  the accuracy  increases. This is because the effect of quantization is similar to adding noise to the activation and regularizing the gradient. But when the number of layers at the client side increases to 5, there is a loss in accuracy, since quantization causes  precision loss in activations and gradients of a large part of the model.

\subsubsection{Communication Reduction}
\begin{figure}[!t]
    \centering
    \includegraphics[scale = 0.5]{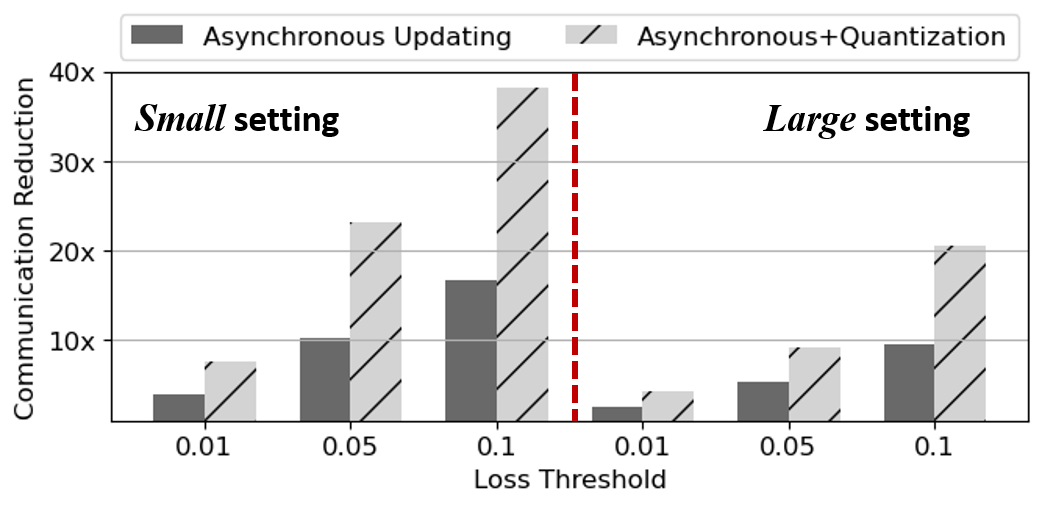}
    \caption{VGG11 Communication reduction with different loss threshold values}
    \label{fig:vgg11-communication}
\end{figure}

\begin{figure}[!t]
    \centering
    \includegraphics[scale = 0.5]{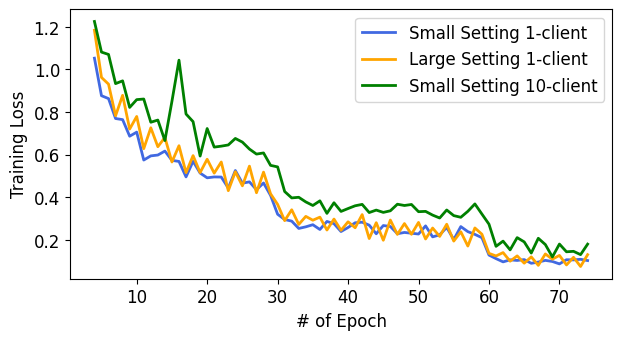}
    \caption{Training loss in split learning of VGG11 for \textit{small} and \textit{large} settings for 1-client and \textit{small} settings for 10-clients during epoch 4 - 75}
    \label{fig:loss_1layer5layers}
\end{figure}

Fig.\ref{fig:vgg11-communication} gives the communication reduction with \textit{small} and \textit{large} client-side setting under different loss threshold values for VGG-11. The communication cost is computed by the bit-volume that has to be transferred between client and server. Higher loss threshold results in fewer updates of client-side model and more significant communication reduction. For the case when only asynchronous training is implemented,  the client-side model is trained only for 8 - 52 epochs (out of 200 epochs) based on the loss threshold; the corresponding communication reduction is 2.6x-16.8x. 
Another interesting result is that, the reduction achieved for \textit{small} setting is better than for \textit{large} setting. This is because the asynchronous training slows down the loss drop of \textit{large} setting during training as shown in Fig.\ref{fig:loss_1layer5layers}.

The communication reduction for VGG11 is even higher at 4.2x- 38.1x when both asynchronous training and quantization are implemented. Compared with the 1.5 hours of communication time for VGG11 shown in Table \ref{tab:spl_comm_comp}, it now takes only less than half an hour.
Since in the search-based quantization the gradient/activation are quantized only when the clipped values are less than 1\%, during training, almost all activations are quantized while only 50\% - 80\% gradients are quantized. Usually, it is the later epochs that are not quantized, which means the gradient of later epochs have higher dynamic range.

\begin{figure}[!t]
    \centering
    \includegraphics[scale=0.5]{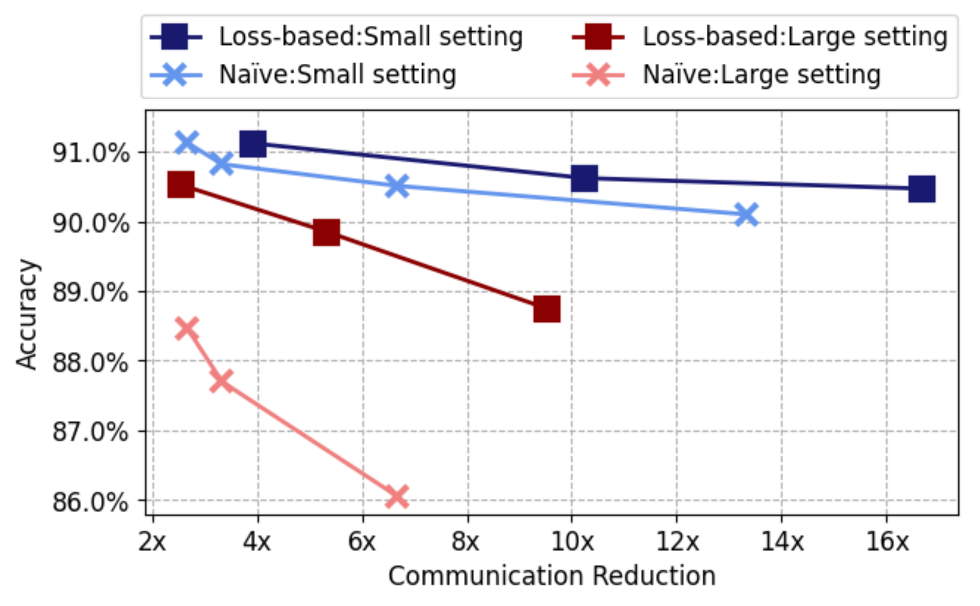}
    \caption{Comparison between loss-based and naïve asynchronous training}
    \label{fig:even_asyncomparison}
\end{figure}

In order to show that our proposed loss-based asynchronous scheme provides a better way of selecting the epoch to update client-side model, we compare it with a naïve asynchronous scheme, where the epoch to update client-side model is uniformly-distributed across all 200 epochs. The accuracy comparison of loss-based and naïve asynchronous method without quantization is shown in Fig.\ref{fig:even_asyncomparison}. Compared to the naïve method, the proposed loss-based method can achieve better accuracy with the same communication reduction. For \textit{small} setting, the proposed loss-based algorithm can achieve slightly better accuracy with the same communication reduction, and for \textit{large} setting, the accuracy of the loss-based method algorithm is more than 1\% higher than the naïve one.

\begin{table}[!h]
\caption{Maximum communication reduction with 0.5\% and 1\% accuracy degradation}
\centering
\resizebox{\linewidth}{!}{
\begin{tabular}{c|c|c|c|c}
\hline
 & \multicolumn{2}{c|}{\textbf{Small Setting}}      & \multicolumn{2}{|c}{\textbf{Large Setting}}                             \\ \cline{2-5} 
                  & \textbf{0.5\% accu. loss} & \textbf{1.0\% accu. loss} & \textbf{0.5\% accu. loss} & \textbf{1.0\% accu. loss} \\ \hline
\textbf{VGG11}  & 23.2x  & 38.1x  &1.64x\protect\footnotemark  & 2.4x \\ 
\hline
\textbf{VGG13}  & 53.33x  & 88.9x & 9.88x & 19.75x \\ 
\hline
\textbf{ResNet18} & 106.7x  & 266.7x & 28.07x & 76.19x \\ 
\hline
\end{tabular}
}
\label{tab:vgg11vgg13resnet}
\end{table}
\addtocounter{footnote}{0}
\footnotetext{Without quantization to keep accuracy loss at less than 0.5\%}
\subsubsection{Other Networks}
Next we present the results of the proposed loss-based asynchronous training and quantization scheme for VGG13 and Resnet18. The baseline accuracies for VGG11, VGG13 and ResNet18 models are 91.5\%, 92.85\% and 94.9\%, respectively. 
Table \ref{tab:vgg11vgg13resnet} reports the maximum communication reduction given 0.5\% and 1\% accuracy degradation for \textit{small} and \textit{large} settings. The maximum communication reduction is achieved when the largest possible loss threshold is chosen for the specified accuracy loss.
Among all three network models, ResNet18 achieves the highest communication reduction implying that the  model updates in ResNet18 are more redundant than others.
The computation reduction is also higher for  ResNet18 since it has fewer model updates.




\subsubsection{Computation Reduction}
The computations at the client side are also reduced due to fewer updates of the client-side model. To approximate the reduction in computation, we use Pytorch built-in profiling function to measure the run-time for activation and gradient computation in clients. The client is modeled by Intel-i7 CPU and the computation without asynchronous training is set as the baseline.
Table \ref{tab:vgg11vgg13resnetcomputation} shows the  computation reduction of clients for VGG11, VGG13 and ResNet18 for 0.5\% and 1.0\% accuracy loss. We see that ResNet18 has the highest  computation reduction (from 7.61x-80.3x) which is expected since ResNet18 has the smallest number of client-model updates.

\begin{table}[!h]
\caption{Computation reduction with 0.5\% and 1\% accuracy degradation}
\centering
\resizebox{\linewidth}{!}{
\begin{tabular}{c|c|c|c|c}
\hline
 & \multicolumn{2}{c|}{\textbf{Small Setting}}      & \multicolumn{2}{|c}{\textbf{Large Setting}}                             \\ \cline{2-5} 
                  & \textbf{0.5\% accu. loss} & \textbf{1.0\% accu. loss} & \textbf{0.5\% accu. loss} & \textbf{1.0\% accu. loss} \\ \hline
\textbf{VGG11}  & 11.8x  & 19.2x  &2.86x  & 5.96x \\ 
\hline
\textbf{VGG13}  & 20.9x  & 27.9x & 2.89x & 8.21x \\ 
\hline
\textbf{ResNet18} & 32.1x  & 80.3x & 7.61x & 20.7x \\ 
\hline
\end{tabular}
}
\label{tab:vgg11vgg13resnetcomputation}
\end{table}

\subsubsection{Multi-client Performance}

As the spirit of distributed learning is to allow multiple clients to collaboratively train a network using their own training samples, next, we present results for the multi-client case. We use the round-robin method to switch epochs among different clients as described in \cite{gupta2018distributed}. Table \ref{tab:multi-client} lists the maximum communication reduction for VGG11 given 0.5\% and 1\% accuracy degradation for \textit{small} setting  when the number of epochs per client is fixed at 200. When the accuracy degradation is 0.5\%, the communication reduction for 5 and 10 clients  is only half compared to that of 1-client case. This is because the drop in loss for the multi-client case is slower than one-client case as shown in Fig.\ref{fig:loss_1layer5layers}, resulting more frequent client-side updates.

\begin{table}[!ht]
\caption{Maximum communication reduction with 0.5\% and 1\% accuracy degradation for multi-client using VGG11}
    \centering

\begin{tabular}{c|c|c}
    \hline
         & \textbf{0.5\% accu. loss} & \textbf{1\% accu. loss }\\\hline
\textbf{1-client }   &23.2x               &38.1x              \\\hline
    \textbf{5-client }   & 11.9x         & 38.9x             \\\hline
\textbf{10-client }   &11.3x                     &41.0x      \\\hline
    \end{tabular}
    \label{tab:multi-client}
\end{table}

\subsubsection{Impact on Privacy}

One of the most significant advantage of split learning over traditional centralized training is  preserving the privacy of user's data. 
Previous work  on privacy of split learning \cite{vepakomma2020nopeek} uses a correlation metric to evaluate the privacy leakage.
The correlation score being more close to 1 means that the output of client-side model is similar to the raw data, implying  that the raw data has a higher chance of being extracted under attack\cite{vepakomma2020nopeek}.  We use the same metric to show the impact on privacy. As shown in Table \ref{tab:privacy}, for VGG11, VGG13 and ResNet18, the privacy for \textit{small} and \textit{large} client-side setting after the proposed scheme is almost intact. So we conclude that the proposed communication reduction method does not affect the privacy of split learning.

\begin{table}[!ht]
\caption{Privacy impact of proposed techniques}
\centering
\begin{tabular}{c|cc|c|c|c}
\hline
 & \multicolumn{2}{c|}{\textbf{Settings}}      & \textbf{VGG11} &\textbf{VGG13} & \textbf{ResNet18}                           \\  \hline
\multirow{3}{*}{\textbf{Small}}  & \multicolumn{2}{c|}{Baseline}  & 0.908 & 0.905 &0.907 \\ 
& \multicolumn{2}{c|}{Only Async}  & 0.891 & 0.894 & 0.949 \\ 
& \multicolumn{2}{c|}{Quant+Async}  & 0.901 & 0.899 & 0.932 \\ 
\hline
\multirow{3}{*}{\textbf{Large}}  & \multicolumn{2}{c|}{Baseline}  & 0.860 & 0.874 &0.834 \\ 
& \multicolumn{2}{c|}{Only Async}  & 0.850 & 0.860 &0.852\\ 
& \multicolumn{2}{c|}{Quant+Async}  & 0.849 & 0.883 &0.856 \\ 
\hline
\end{tabular}
\label{tab:privacy}
\end{table}

\section{Conclusion}
Split learning is a promising privacy-preserving learning scheme that suffers from high communication overhead due to the back and forth passing of activations/gradients  between client and server. In this paper, we propose a loss-based asynchronous training and search-based quantization method for split learning that reduces the communication cost between client and server as well as the computation cost in clients. This is achieved by updating the client-side model only when the loss drop reaches a threshold and by representing the activation/gradient data that is transmitted by 8-bit floating point. The communication reduction methods are validated on VGG11, VGG13 and Resnet18 models using CIFAR10 under various split learning configurations. The results show that for the single-client case, the communication is reduced by 1.64x-106.7x with only 0.5\% accuracy degradation and by 2.4x-266.7x with 1.0\% accuracy degradation. The reduction for 10-client case is smaller at 11.3x and 41.0x for 0.5\% and 1.0\% accuracy loss, respectively.
We also show that the proposed method does not reduce the privacy of user's data compared to the baseline split learning scheme.

\bibliographystyle{./bibliography/IEEEtran}
\bibliography{./bibliography/IEEEabrv}

\end{document}